\documentclass{article}

\usepackage{neurips_2023}

\usepackage[utf8]{inputenc} % allow utf-8 input
\usepackage[T1]{fontenc}    % use 8-bit T1 fonts
\usepackage{hyperref}       % hyperlinks
\usepackage{url}            % simple URL typesetting
\usepackage{booktabs}       % professional-quality tables
\usepackage{amsfonts}       % blackboard math symbols
\usepackage{nicefrac}       % compact symbols for 1/2, etc.
\usepackage{microtype}      % microtypography
\usepackage{xcolor}         % colors
\usepackage{amsmath}        % Include the amsmath package
\usepackage{graphicx}
\usepackage{subcaption, float}

\title{GQKVA: Efficient Pre-training of Transformers by Grouping Queries, Keys, and Values}

\author{Farnoosh Javadi, Walid Ahmed,  Habib Hajimolahoseini,  Foozhan Ataiefard, Mohammad \\ \textbf{Hassanpour, Saina Asani, Austin Wen , Omar Mohamed Awad, Kangling Liu, Yang Liu}\\
$^*$Ascend Team, Toronto Research Center, Huawei Technologies \\
  \texttt {farnoosh.javadi@huawei.com}\\
}

\begin{document}

\maketitle

\begin{abstract}

% %problem
% slow pretraining --> address this by proposing a lighter and faster alternative -->  obtain/observer a trend/linear correlation/ tradeoff between performance and #model_params results to tailored choices baded resource limits -->  offer a fatser and lighter alternatives to MHA --> same perf while less model size. 

% Massive transformer-based models have several challenges including slow and computationally intensive  pre-training and over-parametrization. This paper addresses these issues by proposing a versatile  method called GQKVA,  a generalization of query, key, and value grouping techniques, designed to accelerate transformers pre-training while reducing the model size. Our experiments across various GQKVA variants highlight a clear trade-off between performance and model size, allowing for customized choices based on resource and time constraints. Our findings also reveal that conventional multi-head attention is not always the optimal choice, as lighter and faster alternatives exist. We evaluated our method on ViT which achieved \(\sim0.3\%\) accuracy boost with \(\sim4\%\) less model size on the task of image classification. In addition, our harshest model reduction experiment achieved a reduction of \(\sim\)15\% in model size at the cost of only \(\sim\)1\% accuracy drop.

Massive transformer-based models face several challenges, including slow and computationally intensive pre-training and over-parametrization. This paper addresses these challenges by proposing a versatile method called GQKVA, which generalizes query, key, and value grouping techniques. GQKVA is designed to speed up transformer pre-training while reducing the model size. Our experiments with various GQKVA variants highlight a clear trade-off between performance and model size, allowing for customized choices based on resource and time limitations. Our findings also indicate that the conventional multi-head attention approach is not always the best choice, as there are lighter and faster alternatives available. We tested our method on ViT, which achieved an approximate 0.3\% increase in accuracy while reducing the model size by about 4\% in the task of image classification. Additionally, our most aggressive model reduction experiment resulted in a reduction of approximately 15\% in model size, with only around a 1\% drop in accuracy.

\end{abstract}
\section{Introduction}
Transformers \citep{vaswani2023attention} have dominated RNNs in natural language processing tasks. While CNNs are generally considered the mainstay for various computer vision tasks \citep{wang2023internimage, javadi2020hierarchical, ahmed2023speeding}  transformers have demonstrated their competitive capabilities in many instances  \citep{liu2021swin, touvron2021training, dosovitskiy2020image}. Their scalability and power have prompted a  trend in the literature which is scaling up the transformer-based models by increasing their parameters and layers to achieve superior performance. However, the  expansion of these models has introduced several challenges, such as heavy computation demands and  slow  pre-training, fine-tuning, and inference process.  Furthermore, \citep{hoffmann2022training}  have found out that many of large language models (LLMs) such as GPT3 175B are over-parameterized and inadequately trained, meaning that abundance of parameters doesn't truly translate into enhanced performance. Accordingly, there is a need for techniques that introduce more modestly parameterized transformers to address the issue of over-parametrization. \\
Studies offer various techniques for efficiently fine-tuning massive transformers such as LORA  \citep{hu2021lora, hajimolahoseini2022strategies} and prompt-tuning \citep{lester2021power}. 
 The literature is also rich in the direction of speeding up inference of transformers \citep{ainslie2023gqa, shazeer2019fast, leviathan2023fast, pope2022efficiently}. However, accelerating pre-training of transformer-based models is not well studied. One avenue of research in this direction seeks to lower the time complexity of multi-head attention from being quadratic to linear \citep{choromanski2022rethinking, han2023flatten, wang2020linformer}. This is achieved by introducing a kernel metric that allows for a change in the order of attention computation. However, this direction can lead to accuracy degradation. Another line of research on pre-training acceleration techniques focuses on reducing the number of tokens fed into the transformers \citep{hou2022token, yao2022randomltd}. It is important to note that both of these directions maintain the same model size and do not address the issue of over-parameterization.  This paper aims to contribute to the relatively limited literature that proposes methods for both expediting the pre-training of transformers and reducing the model size simultaneously.
 MQA \citep{shazeer2019fast} and GQA \citep{ainslie2023gqa} have recently shown a promising speedup for speeding up the decoder inference. We propose a more general approach called GQKVA a pre-training acceleration technique. GQKVA partitions queries, keys, and values within the self-attention mechanism to reduce the time needed for attention computation. In our evaluation across vision transformer architecture \citep{dosovitskiy2020image}, we demonstrate that the practice of grouping queries, keys, and values results in expedited training and a more compact model size.  Moreover, we analyzed the trade-offs and implications of grouping Q, K, and Vs in-depth, shedding light on its effects on model convergence, and the  number of parameters.
 Our contributions include: \begin{enumerate}

    \item  Proposing an efficient general attention computation mechanism called GQKVA, which involves grouping queries, keys, and values. 
    \item Thoroughly exploring various ways of grouping Q, K, V matrices during pre-training, including MQA, GQA, MKVA, GKVA, etc.
\item Obtaining a clear trade-off between performance versus model-size and TPS allows for tailored choices based on resource and time constraints.

\end{enumerate} 

% \end{document}

\section{Method}
\label{sec:method}
% In this section, we first recap multi-head attention then will explain other  ways of grouping  queries, keys, and values including MQA, GQA, MKVA, GKVA, and GQKVA.  Figure \ref{fig:model} shows a
% comparison of all the discussed methods. 

\begin{figure}
    \centering
    \includegraphics[width=\textwidth]{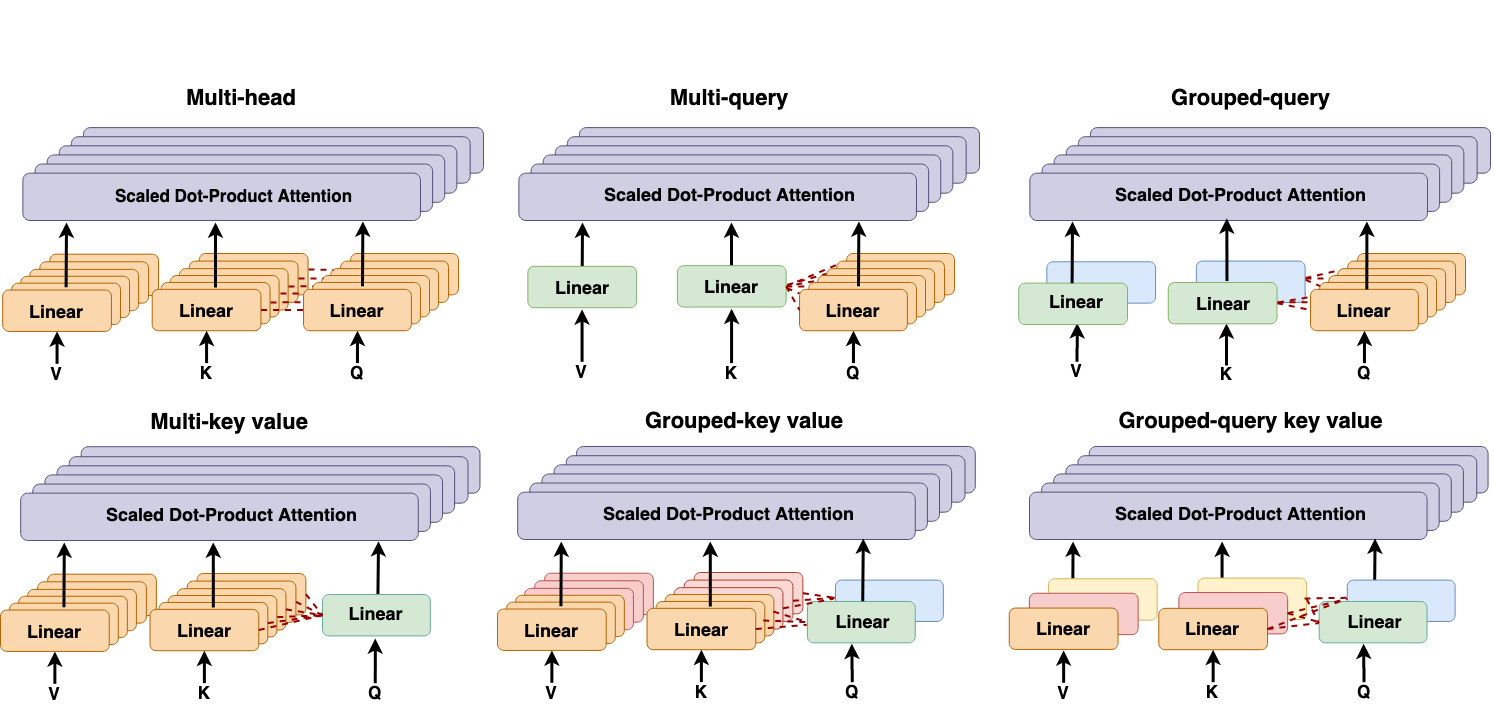} % Replace 'example.png' with your image file name
    \caption {Illustration of various strategies for grouping queries, keys, and values within the attention mechanism, including Vanilla MHA, MQA, GQA, MKVA, GKVA, and GQKVA.}
    \label{fig:model}
\end{figure}
\subsection{Preliminaries}
\textbf{Multi-Head Attention (MHA)} employs a set of \textit{h} distinct attention heads, each ideally specializing in learning unique aspects of the input. For each head, separate query \textit{Q}, key \textit{K}, and value \textit{V} matrices are created by passing the input \textit{x} through a linear layer of dimensions \textit{d} $\times$ \textit{3d}, referred to as the \textbf{qkv layer} (\textit{d} represents the embedding size). The dot-product attention is then computed for each head using Equation \ref{eq:attn}, resulting in unique outputs for each attention head. These output vectors capture diverse aspects of the input and are concatenated before being fed through a second linear layer, which applies a transformation to the combined output.  
\begin{equation}
\text{Attention}(\mathbf{Q}, \mathbf{K}, \mathbf{V}) = \text{softmax}\left(\frac{\mathbf{Q}\mathbf{K}^T}{\sqrt{d}}\right)\mathbf{V}
\label{eq:attn}
\end{equation}
\textbf{Multi-Query Attention (MQA)} initially introduced to enhance inference speed \citep{shazeer2019fast} is a variation of MHA. In this variant, we still maintain \textit{h} distinct heads, each equipped with a separate Q matrix. However, there's a key difference: we employ a single shared K and V matrix across all heads. Consequently, the number of parameters in the qkv layer is reduced to $d \times (d + 2 \times \text{head-dim})$,  $\text{head-dim} = \frac{d}{\text{number of heads}}$. \\

 \textbf{Grouped-Query Attention (GQA)} introduced for faster inference \citep{ainslie2023gqa}, partitions the queries into \textit{g} distinct groups, where each group shares a single \textit{K} and \textit{V}. Consequently, in this scheme, we have \textit{h} Q matrices, along with \textit{g} shared K and V matrices. Notably, when \textit{g} equals 1, GQA is equivalent to MQA, while if \textit{g} equals \textit{h}, GQA aligns with the traditional MHA.  The parameter count for the qkv layer is determined by: $d \times (d + 2\times g \times \text{head-dim})$
 \subsection{Proposed Methods}
 \textbf{Multi-Key Value Attention (MKVA) and Grouped Key Value (GKVA).}
 We propose MKVA and GKVA as novel variations, akin to MQA and GQA. However, MKVA and GKVA differ in that they group keys and values into \textit{g} distinct groups instead of queries; while queries are shared within each group.  It's essential to emphasize that there exists a one-on-one mapping between keys and values. Therefore, however K matrices are grouped, V matrices must undergo the same grouping to maintain this correspondence. To clarify, when a single Q matrix is shared among all \textit{h} Ks and Vs, we refer to the method as MKVA and when \textit{g} Qs are shared, it's called GKVA.\\ 
 
 \textbf{Grouped-Query Key Value Attention (GQKVA)}
To further optimize the parameter count and computational efficiency, we propose a comprehensive approach named GQKVA. In GQKVA, we partition the Q matrices into \textit{$g_q$} groups and the K, V matrices into \textit{$g_{kv}$} groups, where $h = g_q \times g_{kv}$. Subsequently, dot-product attention is computed for each combination of $(Q_i, {KV}_j)_{i \in [1, g_q], j \in [1, g_{kv}] }$ , resulting in \textit{h} distinct outputs similar to the behavior of MHA. It's crucial to note that the usage of Qs and K,Vs must ensure there is no repetition of (Q, KV) pairs to preserve \textit{h} effective heads. Otherwise, we would end up with identical outputs from the dot-product attention, diminishing the model's capacity. GQKVA serves as a unifying generalization of all the methods discussed earlier. For instance, with a single Q matrix shared among all \textit{h} K,Vs, the model corresponds to MKVA. If there are \textit{g} K,Vs shared among all queries, we have GQA-\textit{g}. This approach offers versatility while optimizing parameterization and computational efficiency.
  
\section{Experiments}
\begin{figure}
  \centering
  \begin{subfigure}[b]{0.49\linewidth}
    \includegraphics[width=\linewidth]{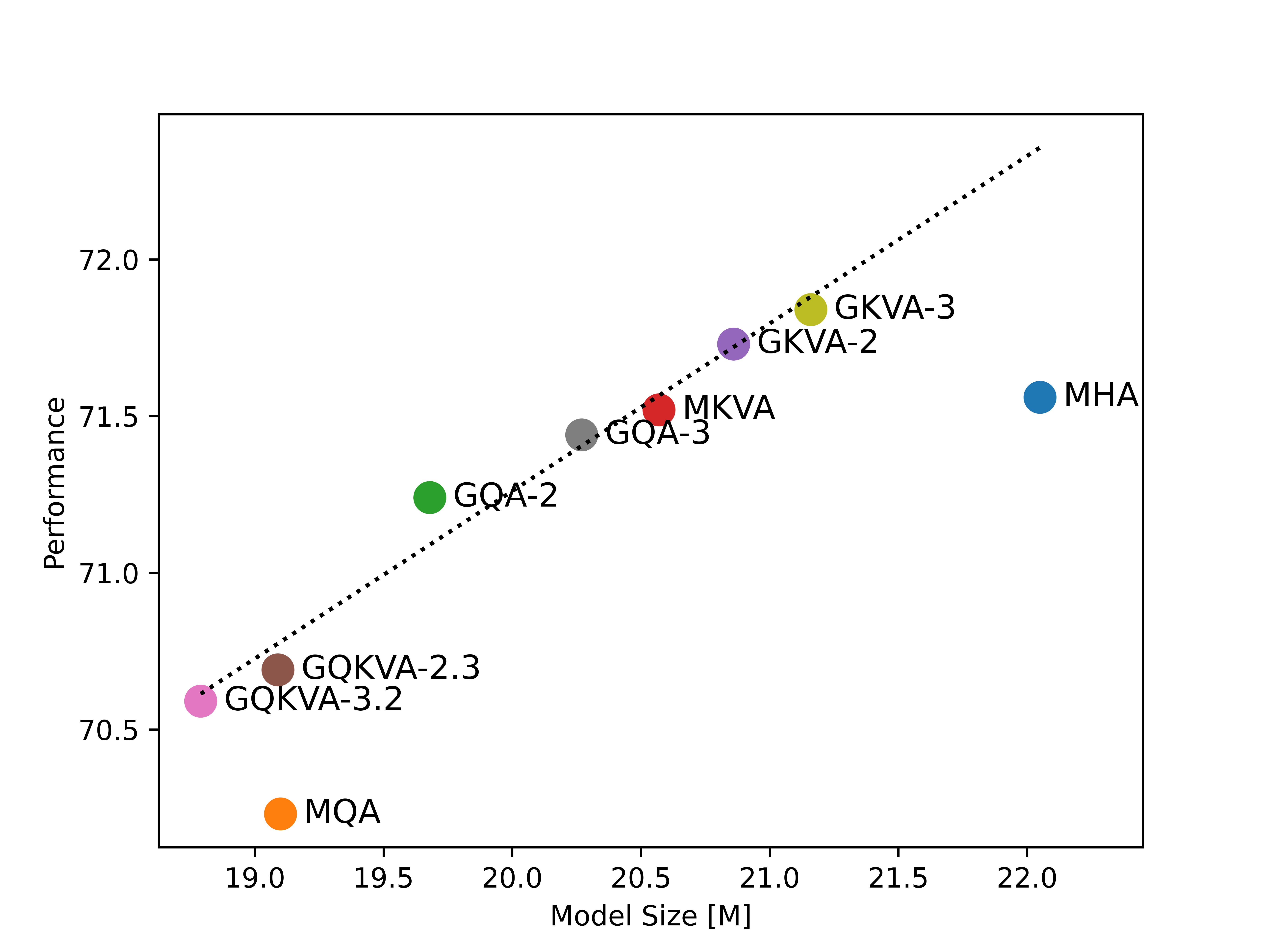}
    \caption{Performance versus Model Size}
    \label{fig:sizeVperf}
  \end{subfigure}
  \begin{subfigure}[b]{0.49\linewidth}
    \includegraphics[width=\linewidth]{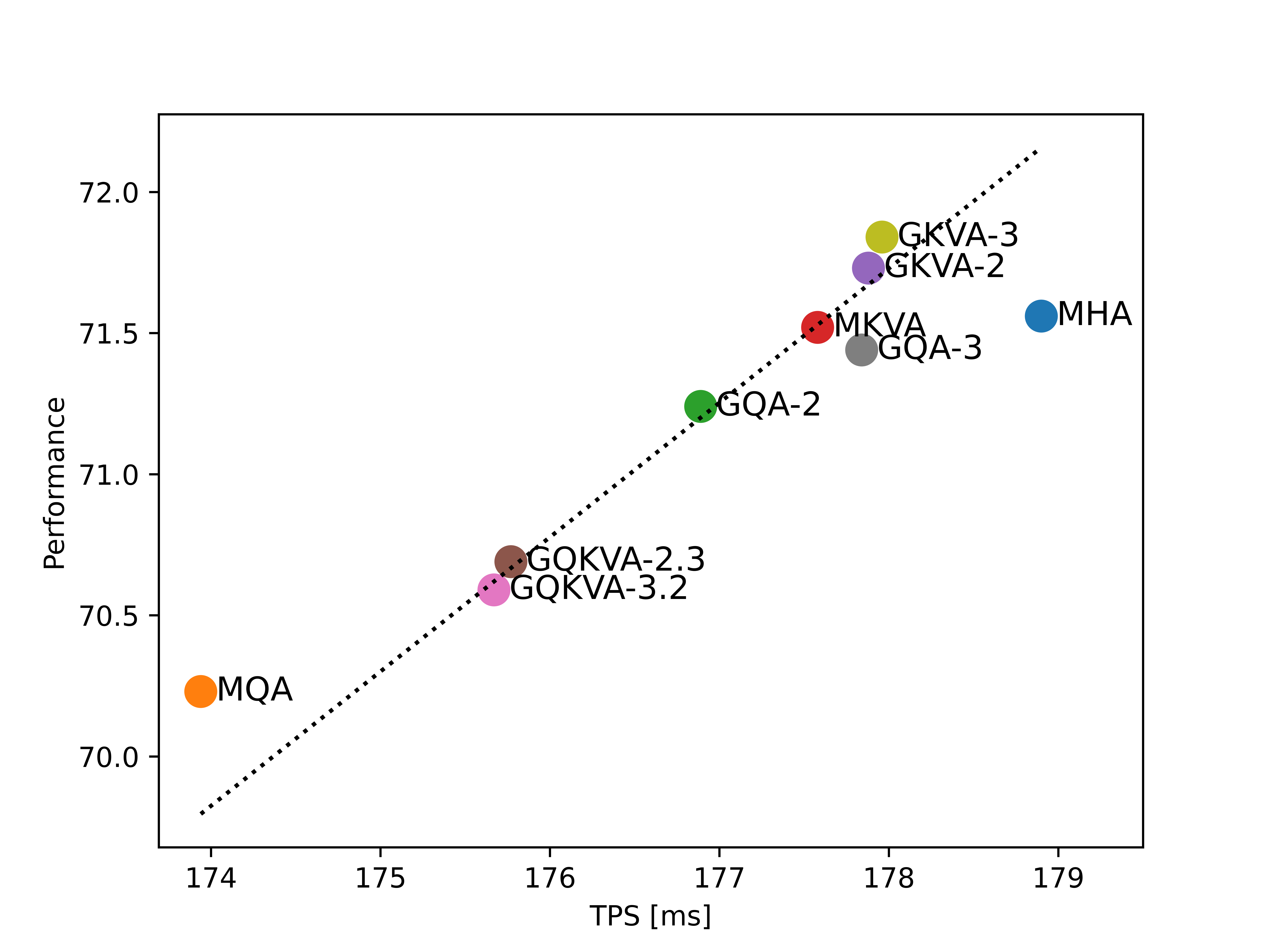}
    \caption{Performance versus TPS}
    \label{fig:timeVperf}
  \end{subfigure}
  \caption{Both figures highlight the presence of faster and lighter attention mechanisms compared to MHA. They also show performance correlates linearly with model size and TPS.}
  \label{fig:perf}
\end{figure}

We assessed the performance of the mentioned methods on ViT \citep{dosovitskiy2020image} for the task of image classification.
We trained ViT-small with 6 heads and 22 million parameters from scratch. The training process was carried out on six 32GB V100 GPU cores using data parallelism for 300 epochs, and a batch size of 288. We used AdamW \citep{adamw} optimizer and an initial learning rate of 0.001.
Through an extensive series of experiments, we compared the following methods: MHA, MQA, GQA-2 (two sets of K,V matrices shared among queries), GQA-3, MKVA, GKVA-2 (two groups of Qs shared among all keys and values), GKVA-3, GQKVA-2.3 (two Q matrices and three K,V matrices), and GQKVA-3.2. The summarized results are presented in Table \ref{table:result}.\\
Our findings indicate that GKVA-2 and GKVA-3 achieved the highest accuracy, reaching 71.73 and 71.84, respectively, while reducing model sizes $5.4\%$ and $4.04\%$ compared to multi-head attention. Moreover, it's seen that GQKVA-2.3 and GQKVA-3.2 have the same or less number of parameters than MQA while outperforming its accuracy respectively by $0.46\%$ and $0.36\%$.
\begin{table}
  \caption{Comparative evaluation of different grouping strategies within the attention mechanism applied on ViT-small as explained in Section \ref{sec:method}. TPS represents the time that training a batch takes. }
  \label{table:result}
  \centering
  \begin{tabular}{|c|c|c|c|c|}
    \hline 
    Model     & TPS        & Acc-top1 & \#Parameters  & Model size  \\
     & (Time per sample - ms) &  (\%) & (M)  &  (mb) \\
   \hline 
    ViT-MHA &  178.90  & 71.56 &22.05 &84.11  \\
    ViT-GKVA-3 & 177.96 (0.53\%)& \textbf{71.84} &21.16 (-4.04\%)& 80.73 \\
    ViT-GKVA-2     &    177.88 (0.57\%)& 71.73&20.86 (-5.40\%) & 79.60   \\
    ViT-MKVA     &   177.58 (0.74\%)& 71.52 &20.57 (-6.71\%) & 78.47  \\
    ViT-GQA-3     &  177.84 (0.59\%) & 71.44 &20.27 (-8.07\%) & 77.34 \\
    ViT-GQA-2    &   176.89 (1.12\%)& 71.24 &19.68 (-10.75\%) & 75.09 \\
    ViT-MQA &   \textbf{173.94} (2.77\%)& 70.23 & 19.09 (-13.42\%) & 72.83\\
    ViT-GQKVA-2.3  &   175.77 (1.75\%)& 70.69&19.09 (-13.42\%) &72.83   \\
    ViT-GQKVA-3.2 &    175.67 (1.82\%)& 70.59&\textbf{18.79 (-14.78\%)}&\textbf{71.70}     \\

    \hline
  \end{tabular}
\end{table}
Figure \ref{fig:sizeVperf}  and \ref{fig:timeVperf} portray the trade-off between model size and performance, as well as the relationship between time per sample (TPS) and performance, respectively. TPS is measured based on the time required to train a batch.  As seen in the figures, both TPS and model size exhibit a linear correlation with performance. Notably, MHA falls below the trend line in both figures, indicating that there are faster and lighter alternatives to MHA and it doesn't necessarily benefit from its larger parameter count.  In general, for methods other than MHA and MQA, an increase in model size leads to improved accuracy, while larger models tend to have higher TPS. Therefore, the choice of the most suitable method can be based on resource and time constraints, allowing for a balanced consideration of model size, training speed, and performance. 

\section{Conclusion}
Transformer models usually suffer from huge model sizes and computationally intensive pre-training. In this paper, we introduce a versatile solution named GQKVA, designed to expedite transformer pre-training while simultaneously reducing model sizes. GQKVA serves as a generalization of various Q, K, V grouping techniques encompassing methods like MQA and GQA. 
% In this method instead of having separate Q,K,V matrices per each head, there are \textit{$g_q$} groups of Q and \textit{$q_{kv}$} groups of K,Vs that will be shared and combined by \textit{h} heads so that overall we still have \textit{h} unique output from each head (\(h = g_k \times g_{kv}\)).
Our experiments involving different GQKVA variants unveil a clear trade-off between model size and performance. This trade-off allows practitioners to tailor their choices based on resource constraints and training time limitations. Our results demonstrate that the conventional MHA is not always the optimal choice, as there exist lighter and faster alternatives.
% The results of eight  experiments training different variants of GQKVA showed a clear trade-off between the model size and performance which one can refer to according to their resource and training time limitations. According to experiments outcome We concluded  that MHA is not always the best choice and there are lighter and faster alternatives for it. 
It's worth noting that the proposed GQKVA method is general and can be applied to any transformer architecture. However, due to our current time and resource constraints, we limited our exploration to the ViT-small model. Future research should extend these techniques to larger transformers, where the potential for greater speed-up and memory savings awaits discovery.

\bibliographystyle{acl_natbib}
\bibliography{refs}

\end{document}